\pdfoutput=1
\documentclass[11pt]{article}

\usepackage{acl}

\usepackage{times}
\usepackage{latexsym}

\usepackage[T1]{fontenc}

\usepackage[utf8]{inputenc}

\usepackage{microtype}

\usepackage{inconsolata}

\usepackage{graphicx}
\usepackage{booktabs}
\usepackage{multirow}
\usepackage{array}
\usepackage{siunitx}
\usepackage[ruled,vlined]{algorithm2e}
\PassOptionsToPackage{table,xcdraw,svgnames,dvipsnames}{xcolor}
\usepackage{colortbl}
\usepackage{dblfloatfix}  
\usepackage{float}
\usepackage{tcolorbox}

\usepackage{pgfplots}
\usepackage{tikz}
\usepackage{amsmath}
\usepackage{diagbox}

\newcommand{\COMMENTLLAMA}[1]{{\textcolor[HTML]{962C38} {$\triangleright$ {#1}\\}}}

\title{DuoDecoding: Hardware-aware Heterogeneous Speculative Decoding\\with Dynamic Multi-Sequence Drafting
}

\author{Kai Lv\textsuperscript{1}\thanks{Work done during internship at Shanghai AI Laboratory.}, Honglin Guo\textsuperscript{1}, Qipeng Guo\textsuperscript{2}, Xipeng Qiu\textsuperscript{1}\\
\textsuperscript{1}Fudan University \  \textsuperscript{2}Shanghai AI Laboratory\\
klv23@m.fudan.edu.cn}

\begin{document}
\maketitle
\begin{abstract}
Large language models (LLMs) exhibit exceptional performance across a wide range of tasks; however, their token-by-token autoregressive generation process significantly hinders inference speed. 
Speculative decoding presents a promising draft-then-verify framework that reduces generation latency while maintaining output distribution fidelity.
Nevertheless, the draft model introduces additional computational overhead, becoming a performance bottleneck and increasing the time to first token (TTFT). 
Previous approaches to mitigate draft model overhead have primarily relied on heuristics and generally failed to match the quality of the draft language models.
To address these challenges, we propose DuoDecoding, a novel approach that strategically deploys the draft and target models on the CPU and GPU respectively, enabling parallel decoding while preserving draft quality.
Our method incorporates a hardware-aware optimal draft budget to minimize idle times and employs dynamic multi-sequence drafting to enhance draft quality. 
Extensive experiments across seven tasks show that DuoDecoding achieves up to 2.61x speedup in generation latency, while reducing TTFT to 83\% of that in conventional speculative decoding. The Code is available at \url{https://github.com/KaiLv69/DuoDecoding}.
\end{abstract}

\section{Introduction}

Large language models (LLMs) have demonstrated impressive performance across a wide range of domains and have been extensively deployed~\cite{gpt4,llama3,qwen25,deepseek2,internlm2}. However, their massive parameter sizes and computational requirements pose significant deployment challenges. In particular, the autoregressive generation process~\cite{transformer} requires a complete forward pass of the entire model for each new token sequentially, leading to considerable latency and limiting practical utility.

Recent advances in speculative decoding~\cite{chen2025sequoia,sun2024triforce,EAGLE,EAGLE-2,GliDeCaPE} have shown promise in reducing latency without compromising generation quality.
Speculative decoding treats the original language model as the target model and employs a smaller draft model to speculate the target model's output.
Each decoding iteration consists of two phases: (1) the draft phase, where the draft model autoregressively generates multiple candidate tokens, and (2) the verification phase, where the target model evaluates all candidate tokens in a single forward pass and verifies them through speculative sampling according to the output distribution. This process allows the target model to generate multiple tokens in a single forward pass while maintaining its original output distribution~\cite{leviathan2023fast}.

\begin{figure}[t]
    \centering
    \includegraphics[width=1\linewidth]{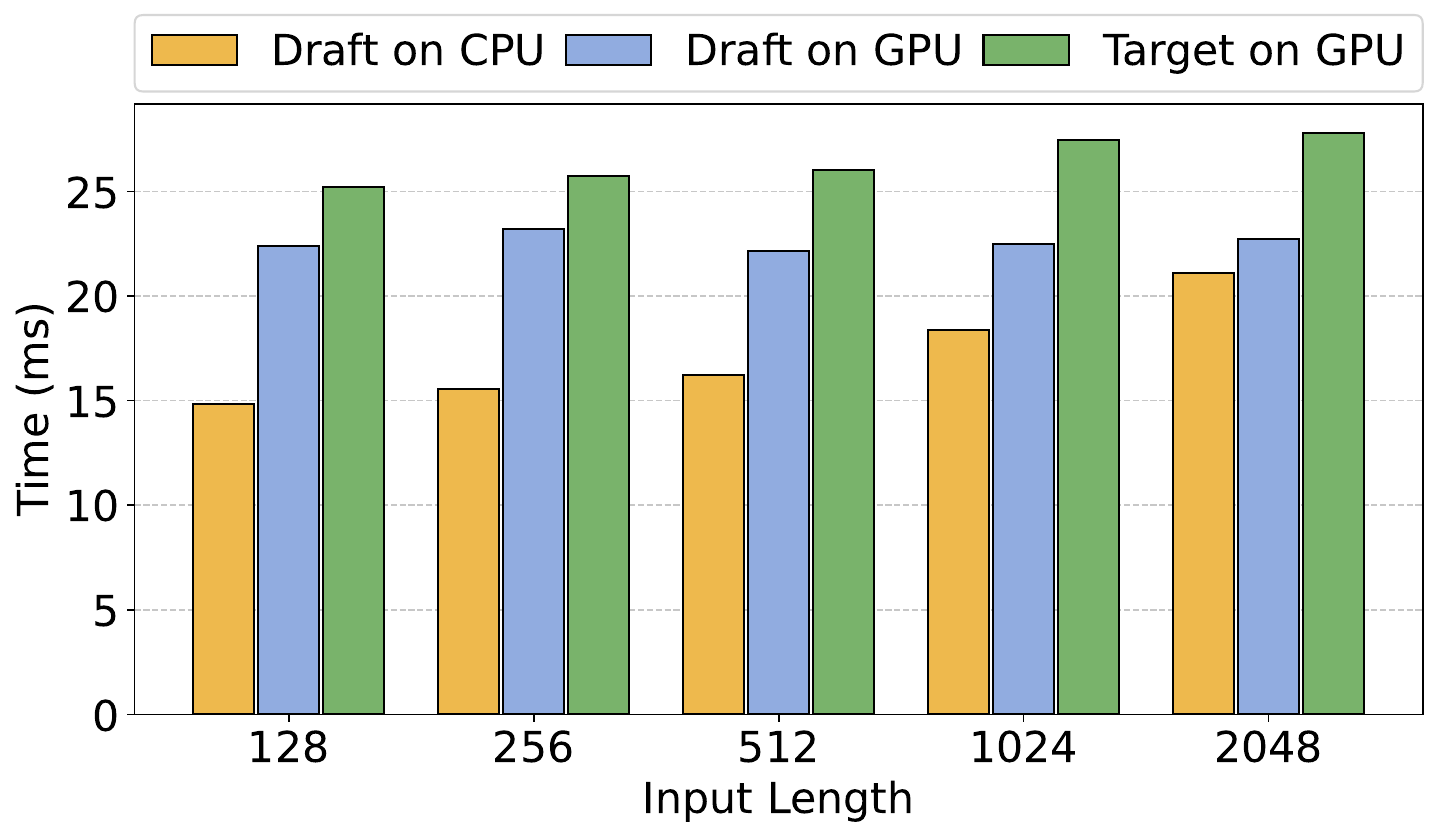}
    \caption{Wall time for draft model autoregressive generation and target model parallel verification of 8 tokens with varying input lengths. The draft phase has become a comparable bottleneck to the verification phase, and executing the lightweight draft model on CPU does not compromise generation efficiency.}
    \label{fig:time}
\end{figure}

However, the draft model introduces additional computational overhead and the draft phase has emerged as a bottleneck comparable to the verification phase~\cite{zafrir2024fastdraft,anonymous2024designing,pearl}.
The draft model also introduces undesirable side effects, including increased GPU memory consumption and longer time to first token (TTFT).
While several approaches have been proposed to reduce the draft model overhead, they generally rely on heuristics, failing to match the draft quality of draft language models~\cite{pld,Lookahead,REST,zhao2024ouroboros}.

In contrast to previous approaches, we propose deploying the draft model on CPU, which shifts additional computational overhead to CPU without compromising draft quality. 
A key assumption underlying this approach is that the draft model should maintain an acceptable generation speed on CPU; otherwise, the potential  acceleration effect would be limited.
We provide empirical validation in Figure~\ref{fig:time}, which demonstrates that the wall time required for auto-regressive generation of 8 tokens by the draft model (\texttt{Llama-68m}) on CPU matches that of the parallel verification of 8 tokens by the target model (\texttt{Llama-2-7B}) on GPU across various sequence lengths.

In this paper, we introduce DuoDecoding, a hardware-aware heterogeneous speculative decoding method with dynamic multi-sequence drafting.
By deploying the draft model and target model on CPU and GPU respectively, we not only shift the overhead of draft model to CPU, but also enable concurrent execution of the draft and verification processes.
During this parallel decoding process, we employ a hardware-aware optimal draft budget to minimize idle time on either CPU or GPU.
When this draft budget becomes high, tokens positioned later in the draft sequence exhibit diminished acceptance rates.
To improve the acceptance rate of the draft, we introduce dynamic multi-sequence drafting based on the uncertainty of draft outputs.
Additionally, we adapt the verification procedure for our novel speculative decoding process, ensuring that the output distribution of DuoDecoding is consistent with that of the target model.

Experimental results across seven different tasks demonstrate that DuoDecoding can significantly reduce the generation latency of LLMs, achieving up to a 2.61x speedup. 
Compared to conventional speculative decoding~\cite{chen2023accelerating,leviathan2023fast}, DuoDecoding achieves a 17\% reduction in time to first token (TTFT). 
Comprehensive ablation studies demonstrate the contribution of each component in our proposed method.
Through detailed analysis, we validate the effectiveness of our uncertainty-based dynamic multi-sequence drafting strategy.

\section{Related Work}
\paragraph{Speculative Decoding}
\citet{stern2018blockwise} introduces a draft-then-verify generation framework to improve the generation speed of autoregressive models through increasing parallelism. 
The verification phase of speculative decoding~\cite{chen2025sequoia,sun2024triforce} is centered around speculative sampling~\cite{leviathan2023fast,chen2023accelerating}. By comparing vocabulary-level probabilities between the draft and target models, this technique achieves higher acceptance rates than conventional rejection sampling while maintaining consistency with the target model's output distribution.
BiLD~\cite{kim2024speculative}, Medusa~\cite{Medusa}, and Hydra~\cite{ankner2024hydra} propose different modified verification strategies, exploring the trade-off between generation speed and distribution fidelity.
While our method also incorporates a draft-verify mechanism, it innovates by executing draft and target models concurrently rather than sequentially.

\paragraph{Draft Overhead Reduction}
The additional overhead introduced by the draft step has become one of the main bottlenecks in speculative decoding~\cite{zafrir2024fastdraft,anonymous2024designing}, leading to a series of efforts aiming to address this challenge. Lookahead Decoding~\cite{Lookahead} caches n-grams generated during decoding and employs Jacobian decoding~\cite{jacobi} for parallel generation. 
Ouroboros~\cite{zhao2024ouroboros} constructs a list of phrases to accelerate the draft models and lengthen the drafts.
PLD~\cite{pld} retrieves n-grams from the prompt to serve as drafts. REST~\cite{REST} extends this idea by establishing a larger-scale retrieval corpus and using longest prefix matching from the datastore to generate drafts. However, these methods generally yield outputs with lower distribution alignment to the target model compared to dedicated draft models.
Most similar to our work, \citet{pearl} utilizes additional GPU resources to distribute the draft overhead. Our work differs by identifying and exploiting the potential of heterogeneous devices, while dynamically adapting the draft process based on available computational resources and draft outputs.

\begin{algorithm*}[ht]
\caption{DuoDecoding}
\label{algo:duodec}
\KwIn{target model $M_p$, draft model $M_q$, prefix $\mathbf{x}=[x_1, x_2, \dotsc,x_{n}]$, max generation tokens $L$, hardware-aware optimal draft budget $\gamma$}

\textbf{Initialize:} empty unverified prefix $\mathbf{\tilde{x}}$

\texttt{init\_process\_group(world\_size=2)}

\While{$n-\mathbf{\tilde{x}}$.\texttt{length} < L}{
\COMMENTLLAMA{Forward in parallel on heterogeneous devices}

\begin{tabular}{@{}l  l@{}}
    \textbf{Draft Process on CPU:} & \textbf{Target Process on GPU:} \\
   \quad$\mathbf{q}_{\leq n}, \mathbf{\hat{q}}_{[s, s/\gamma]}, \mathbf{\hat{x}}_{[s,s/\gamma]} \gets$ \texttt{dynamic\_drafting($\mathbf{x}_{\leq n}$, $\gamma$)} & \quad $\mathbf{p}_{\leq n} \gets M_p(\mathbf{x}_{\leq n})$
\end{tabular}
\vspace{3pt}\\
\COMMENTLLAMA{Inter-process communication}
Synchronization probability via inter-process communication\\

\COMMENTLLAMA{Verification (Details in Algorithm~\ref{algo:verify})}
$n, \mathbf{x}, \mathbf{\tilde{x}} \gets$  \texttt{DuoDecVerify($n, \mathbf{x}, \mathbf{\tilde{x}},\mathbf{q}_{\leq n}, \mathbf{\hat{q}}_{[s, s/\gamma]}, \mathbf{\hat{x}}_{[s,s/\gamma]},\mathbf{p}_{\leq n}$)}

}

\end{algorithm*}
\paragraph{Draft Performance Enhancement}
The draft outputs directly influence the acceptance rate. 
Typically, improving the alignment between the output distributions of the draft model and the target model enhances draft performance.
DistillSpec~\cite{DistillSpec} and Online Speculative Decoding~\cite{OnlineSD} employ knowledge distillation techniques to enhance distribution consistency.
Glide~\cite{GliDeCaPE} improves performance by training draft models to reuse the target model's KV cache.
Eagle~\cite{EAGLE,EAGLE-2} introduces an additional layer in the target model to perform autoregressive generation at the feature level, thereby improving prediction accuracy.
SpecInfer~\cite{miao2023specinfer} combines multiple draft models, each fine-tuned collectively, to jointly predict the outputs of target model.
These approaches are orthogonal to our method and could potentially be integrated with our framework for complementary benefits.

\section{Method}
As shown in Algorithm~\ref{algo:duodec}, the overall process of DuoDecoding can be divided into three stages: parallel execution of the draft model and target model on heterogeneous devices, communication for synchronizing output probabilities, and verification.

\subsection{Heterogeneous Parallel Decoding}
The distinct computational requirements of the target and draft models in speculative decoding naturally lend themselves to deployment across heterogeneous computing devices. 
For typical server setups, the target model can be placed on the GPU, while the computationally lighter draft model can run on the CPU.
This heterogeneous deployment strategy not only alleviates the computational burden on the GPU but also enables concurrent execution of both models. Consequently, it eliminates the sequential dependencies inherent in traditional speculative decoding, enhancing parallelism and addressing the performance bottleneck caused by the draft model's overhead.

Specifically, in each iteration, both the draft and target models receive identical inputs and run simultaneously. The draft model, running on CPU, autoregressively generates multiple tokens to speculate the target model's output. Concurrently, the target model, executing on GPU, validates all draft tokens from the previous iteration and predicts the next token, which serves to verify the draft results in the current iteration.

\paragraph{Hardware-aware Drafting Budget}
In parallel decoding, optimal performance requires balancing the execution time between CPU and GPU operations. 
Inefficiencies arise when CPU processing either finishes prematurely, suggesting potential for increased drafting length, or extends too long, leading to idle time on the GPU.

The optimal drafting budget varies across different hardware configurations due to varying relative speeds between GPU and CPU. 
We propose to measure the cost coefficient $c$, defined as the ratio of forward pass time between the target model on GPU and draft model on CPU. 
By setting the drafting budget $\gamma$ equal to $c$, we achieve approximate temporal alignment between draft and target model executions, thereby maximizing hardware utilization efficiency.

\subsection{Dynamic Multi-Sequence Drafting}
As the drafting budget increases, the acceptance rate for tokens in later positions of a single drafted sequence tends to decline significantly. Given that each iteration validates at most the first token generated in the current draft, we propose multi-sequence drafting to maximize the utilization of this validation information.

\paragraph{Draft Uncertainty as Proxy}
Since we cannot know in advance whether a draft token will be accepted, we use the predicted probabilities from the draft model as a proxy for acceptance rates. Intuitively, higher draft probabilities indicate greater model confidence at that position, suggesting a higher likelihood of correct prediction.

\begin{figure}[t]
    \centering
    \includegraphics[width=1\linewidth]{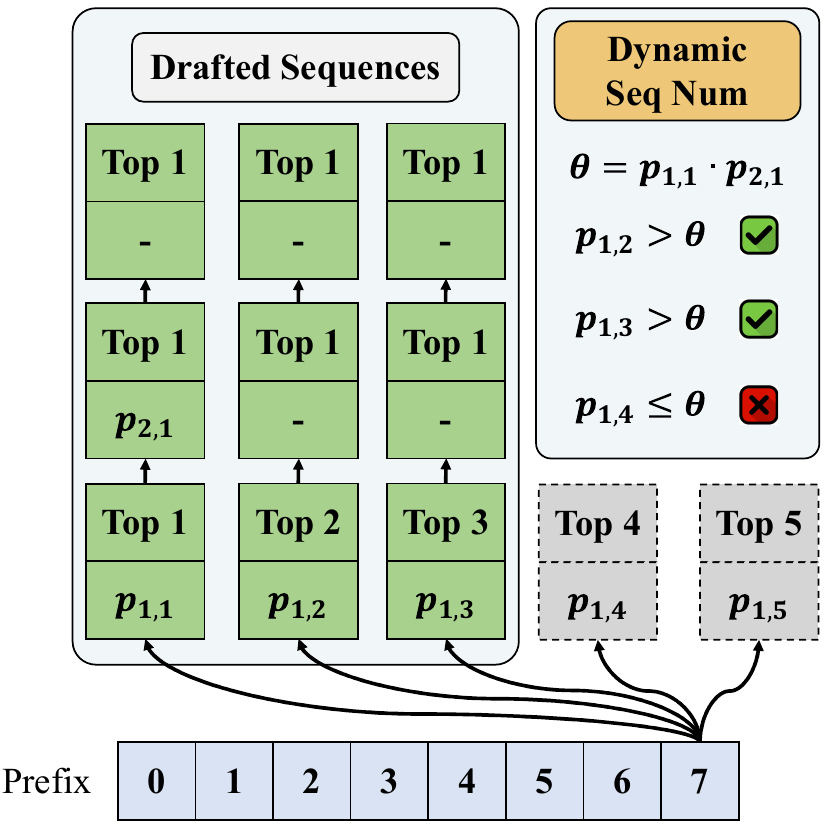}
    \caption{Dynamic multi-sequence drafting. $p_{i,j}$ represents the probability of the $j$-th ranked token at the $i$-th position in the generated sequence. $\theta= p_{1,1} \times p_{2,1}$ serves as the threshold.
    Tokens with probabilities $p_{1,k}$ exceeding the threshold $\theta$ will continue to be predicted sequentially, forming a independent draft sequence.
    }
    \label{fig:dynamic-seq}
\end{figure}

\paragraph{Draft Sequences Construction}
The number of draft sequences is dynamically determined based on the draft uncertainty.
Figure~\ref{fig:dynamic-seq} illustrates the construction process of draft sequences. Let $p_{i,j}$ denote the probability of the $j$-th ranked token at position $i$ in the generation sequence. We use $p_{i,j}$ to approximate the acceptance probability of this token. The threshold $\theta = p_{1,1} \times p_{2,1}$ represents the probability of accepting the first two tokens in the sequence with the highest probabilities. To improve acceptance rates in subsequent drafting, we search for tokens at the first position of the generation sequence, whose acceptance rate exceeds the threshold $\theta$ to generate the next token. Specifically, tokens with $p_{1,k}$ whose probabilities exceed $\theta$ will continue to be predicted sequentially to form an independent sequence.

\begin{algorithm}[t]
\caption{Verification Process}
\label{algo:verify}
\KwIn{$n, \mathbf{x}, \mathbf{\tilde{x}},\mathbf{q}_{\leq n}, \mathbf{\hat{q}}_{[s, s/\gamma]}, \mathbf{\hat{x}}_{[s,s/\gamma]},\mathbf{p}_{\leq n}$ from Algorithm~\ref{algo:duodec}}

\SetKwProg{Fn}{Function}{:}{\KwRet}
  \Fn{\texttt{DuoDecVerify}}{
    \COMMENTLLAMA{Verify the unverified part in the prefix}
    $l\gets \mathbf{\tilde{x}}.\texttt{length}$ \\
    $r_1, \dots, r_l \sim \texttt{Uniform}(0,1)$ \\
    $k\gets \texttt{max}(\{i|0\leq i < l, r_{n-i} > \frac{p_{n-i}}{q_{n-i}}\})$ \\


    \eIf{$\mathbf{\tilde{x}}.\texttt{length} == 0$ \texttt{or} $k == 0$}
    {   
        \COMMENTLLAMA{Verify multi-sequence drafts}

        $k\gets -1$ \\
        \For{$i\gets1$ \KwTo $s$}{
            $r_i \sim \texttt{Uniform}(0, 1)$\\
            \eIf{$r_i < p_n / \hat{q}_{i,0}$}{
                $k \gets i$ \\
                $\mathbf{x} \gets [x_1, x_2, \dots, x_n] + \mathbf{\hat{x}}_{k,:} $ \\
                $n\gets n+k/\gamma$ \\
                $\mathbf{\tilde{x}} \gets \hat{\mathbf{x}}_{s,:} $ \\
                \KwRet $n, \mathbf{x}, \mathbf{\tilde{x}}$
            }{
                $p'_n \gets \texttt{norm}(\texttt{max}(p_{n}-\hat{q}_{i,0},0))$
            }
        }
        
        \COMMENTLLAMA{All sequences rejected}
        
        \If{$k == -1$}
        {
        $t\sim p'_{n}$ \\
        $\mathbf{x} \gets [x_1,x_2,\dots,x_{n},t]$ \\
        $n\gets n+1$\\
        empty $\mathbf{\tilde{x}}$\\
        }
    }
    {
        $t\sim \texttt{norm}(\texttt{max}(p_{n-k}-q_{n-k},0))$ \\
        $\mathbf{x} \gets [x_1,x_2,\dots,x_{n-k},t]$ \\
        $n\gets n-k+1$\\
        empty $\mathbf{\tilde{x}}$\\
    }
    \KwRet $n, \mathbf{x}, \mathbf{\tilde{x}}$
  }

\end{algorithm}

\subsection{Verification}
\begin{table*}[ht]
\centering
\footnotesize
\resizebox{\textwidth}{!}{
\begin{tabular}{@{} c@{} l@{} S[table-format=3.2]@{} S[table-format=1.2]@{} S[table-format=3.2]@{} S[table-format=1.2]@{} S[table-format=3.2]@{} S[table-format=1.2]@{} S[table-format=3.2]@{} S[table-format=1.2]@{} S[table-format=3.2]@{} S[table-format=1.2]@{} S[table-format=3.2]@{} S[table-format=1.2]@{} S[table-format=3.2]@{} S[table-format=1.2]@{} S[table-format=3.2] @{~~} S[table-format=1.2]@{}}

\toprule
\multicolumn{2}{l}{\multirow{2}{*}{\textbf{Method}}} &
  \multicolumn{2}{c}{\textbf{MT-Bench}} &
  \multicolumn{2}{c}{\textbf{Trans}} &
  \multicolumn{2}{c}{\textbf{Sum}} &
  \multicolumn{2}{c}{\textbf{QA}} &
  \multicolumn{2}{c}{\textbf{Math}} &
  \multicolumn{2}{c}{\textbf{RAG}} &
  \multicolumn{2}{c}{\textbf{Code}} &
  \multicolumn{2}{c}{\textbf{Avg.}} \\ \cmidrule(lr){3-4} \cmidrule(lr){5-6} \cmidrule(lr){7-8}  \cmidrule(lr){9-10} \cmidrule(lr){11-12} \cmidrule(lr){13-14} \cmidrule(lr){15-16} \cmidrule(lr){17-18}
 &
   &
  \textit{TPS} &
  $\varphi$ &
  \textit{TPS} &
  $\varphi$ &
  \textit{TPS} &
  $\varphi$ &
  \textit{TPS} &
  $\varphi$ &
  \textit{TPS} &
  $\varphi$ &
  \textit{TPS} &
  $\varphi$ &
  \textit{TPS} &
  $\varphi$ &
  \textit{TPS} &
  $\varphi$ \\ \midrule
\multirow{6}{*}{\rotatebox{90}{Vicuna}} &
  \multicolumn{1}{l|}{Vanilla} &
  44.78 &
  \multicolumn{1}{c|}{1.00} &
  45.62 &
  \multicolumn{1}{c|}{1.00} &
  44.29 &
  \multicolumn{1}{c|}{1.00} &
  43.78 &
  \multicolumn{1}{c|}{1.00} &
  44.65 &
  \multicolumn{1}{c|}{1.00} &
  44.03 &
  \multicolumn{1}{c|}{1.00} &
  45.10 &
  \multicolumn{1}{c|}{1.00} &
  44.61 &
  1.00 \\
 &
  \multicolumn{1}{l|}{SpS} &
  63.65 &
  \multicolumn{1}{c|}{1.42} &
  56.74 &
  \multicolumn{1}{c|}{1.24} &
  66.77 &
  \multicolumn{1}{c|}{1.51} &
  60.08 &
  \multicolumn{1}{c|}{1.37} &
  63.38 &
  \multicolumn{1}{c|}{1.42} &
  66.50 &
  \multicolumn{1}{c|}{1.51} &
  67.57 &
  \multicolumn{1}{c|}{1.50} &
  63.53 &
  1.42 \\
 &
  \multicolumn{1}{l|}{PLD} &
  62.66 &
  \multicolumn{1}{c|}{1.40} &
  55.14 &
  \multicolumn{1}{c|}{1.26} &
  \textbf{\ \ 95.31} &
  \multicolumn{1}{c|}{\textbf{2.15}} &
  48.74 &
  \multicolumn{1}{c|}{1.11} &
  66.03 &
  \multicolumn{1}{c|}{1.48} &
  \textbf{\ \ 75.21} &
  \multicolumn{1}{c|}{\textbf{1.71}} &
  55.81 &
  \multicolumn{1}{c|}{1.24} &
  65.56 &
  1.47 \\
 &
  \multicolumn{1}{l|}{REST} &
  58.83 &
  \multicolumn{1}{c|}{1.31} &
  49.85 &
  \multicolumn{1}{c|}{1.09} &
  54.52 &
  \multicolumn{1}{c|}{1.23} &
  \textbf{\ \ 65.82} &
  \multicolumn{1}{c|}{\textbf{1.50}} &
  50.98 &
  \multicolumn{1}{c|}{1.14} &
  60.76 &
  \multicolumn{1}{c|}{1.38} &
  69.93 &
  \multicolumn{1}{c|}{1.55} &
  58.67 &
  1.32 \\
 &
  \multicolumn{1}{l|}{Lookahead} &
  62.18 &
  \multicolumn{1}{c|}{1.39} &
  62.57 &
  \multicolumn{1}{c|}{1.37} &
  58.72 &
  \multicolumn{1}{c|}{1.33} &
  57.67 &
  \multicolumn{1}{c|}{1.32} &
  66.70 &
  \multicolumn{1}{c|}{1.49} &
  55.81 &
  \multicolumn{1}{c|}{1.27} &
  59.34 &
  \multicolumn{1}{c|}{1.32} &
  60.43 &
  1.35 \\
 &
  \multicolumn{1}{l|}{DuoDec} &
  \textbf{\ \ 74.73} &
  \multicolumn{1}{c|}{\textbf{1.67}} &
  \textbf{\ \ 66.02} &
  \multicolumn{1}{c|}{\textbf{1.45}} &
  73.76 &
  \multicolumn{1}{c|}{1.67} &
  65.38 &
  \multicolumn{1}{c|}{1.49} &
  \textbf{\ \ 68.54} &
  \multicolumn{1}{c|}{\textbf{1.54}} &
  73.12 &
  \multicolumn{1}{c|}{1.66} &
  \textbf{\ \ 72.00} &
  \multicolumn{1}{c|}{\textbf{1.60}} &
  \textbf{\ \ 70.51} &
  \textbf{1.58} \\ \midrule
\multirow{6}{*}{\rotatebox{90}{Llama}} &
  \multicolumn{1}{l|}{Vanilla} &
  44.31 &
  \multicolumn{1}{c|}{1.00} &
  44.10 &
  \multicolumn{1}{c|}{1.00} &
  44.22 &
  \multicolumn{1}{c|}{1.00} &
  43.87 &
  \multicolumn{1}{c|}{1.00} &
  44.98 &
  \multicolumn{1}{c|}{1.00} &
  39.99 &
  \multicolumn{1}{c|}{1.00} &
  44.76 &
  \multicolumn{1}{c|}{1.00} &
  43.75 &
  1.00 \\
 &
  \multicolumn{1}{l|}{SpS} &
  88.01 &
  \multicolumn{1}{c|}{1.99} &
  96.39 &
  \multicolumn{1}{c|}{2.19} &
  78.65 &
  \multicolumn{1}{c|}{1.78} &
  56.94 &
  \multicolumn{1}{c|}{1.30} &
  105.63 &
  \multicolumn{1}{c|}{2.35} &
  49.73 &
  \multicolumn{1}{c|}{1.24} &
  76.57 &
  \multicolumn{1}{c|}{1.71} &
  78.85 &
  1.80 \\
 &
  \multicolumn{1}{l|}{PLD} &
  86.18 &
  \multicolumn{1}{c|}{1.94} &
  100.69 &
  \multicolumn{1}{c|}{2.28} &
  \textbf{134.97} &
  \multicolumn{1}{c|}{\textbf{3.05}} &
  52.29 &
  \multicolumn{1}{c|}{1.19} &
  106.28 &
  \multicolumn{1}{c|}{2.36} &
  79.84 &
  \multicolumn{1}{c|}{2.00} &
  83.00 &
  \multicolumn{1}{c|}{1.85} &
  91.89 &
  2.10 \\
 &
  \multicolumn{1}{l|}{REST} &
  62.17 &
  \multicolumn{1}{c|}{1.40} &
  55.53 &
  \multicolumn{1}{c|}{1.26} &
  54.32 &
  \multicolumn{1}{c|}{1.23} &
  64.45 &
  \multicolumn{1}{c|}{1.47} &
  56.98 &
  \multicolumn{1}{c|}{1.27} &
  53.31 &
  \multicolumn{1}{c|}{1.33} &
  73.41 &
  \multicolumn{1}{c|}{1.64} &
  60.02 &
  1.37 \\
 &
  \multicolumn{1}{l|}{Lookahead} &
  73.54 &
  \multicolumn{1}{c|}{1.66} &
  77.74 &
  \multicolumn{1}{c|}{1.76} &
  61.88 &
  \multicolumn{1}{c|}{1.40} &
  47.48 &
  \multicolumn{1}{c|}{1.08} &
  85.22 &
  \multicolumn{1}{c|}{1.89} &
  45.49 &
  \multicolumn{1}{c|}{1.14} &
  68.33 &
  \multicolumn{1}{c|}{1.53} &
  65.67 &
  1.50 \\ 
  &
  \multicolumn{1}{l|}{DuoDec} &
  \textbf{101.67} &
  \multicolumn{1}{c|}{\textbf{2.29}} &
  \textbf{139.08} &
  \multicolumn{1}{c|}{3.15} &
  85.84 &
  \multicolumn{1}{c|}{1.94} &
  \textbf{139.57} &
  \multicolumn{1}{c|}{\textbf{3.18}} &
  \textbf{150.67} &
  \multicolumn{1}{c|}{\textbf{3.35}} &
  \textbf{\ \ 92.58} &
  \multicolumn{1}{c|}{\textbf{2.32}} &
  \textbf{\ \ 89.52} &
  \multicolumn{1}{c|}{\textbf{2.00}} &
  \textbf{114.13} &
  \textbf{2.61} \\ \bottomrule
\end{tabular}
}
\caption{Performance comparison across different tasks and models. We report tokens per second (\textit{TPS}) and speedup ratio ($\varphi$) relative to vanilla autoregressive generation for both \texttt{Vicuna-7b-v1.5} and \texttt{Llama2-7b} models. Higher values indicate better performance. The best results are highlighted in bold.}
\label{tab:main-results}
\end{table*}
We design our verification strategy based on speculative sampling~\cite{leviathan2023fast,chen2023accelerating}. Since the draft and target models receive identical inputs in each generation iteration, some draft tokens may not be verified within the same iteration they are generated. Therefore, we first verify the draft tokens that were not verified in the previous iteration. This verification process follows the same procedure as speculative sampling.

If all draft tokens from the previous iteration have been successfully verified, we proceed with the verification of the first token in the multi-sequence draft. Verified sequences are then appended to the prefix for the next iteration. If none of the sequences are verified, we sample a token from the normalized distribution and append it to the prefix for the next iteration.

\section{Experiment}

\subsection{Setup}

\paragraph{Tasks} To comprehensively evaluate our method's effectiveness across different scenarios, we conduct experiments on seven diverse task categories. We incorporate the widely-adopted SpecBench~\cite{specbench} and extend our evaluation to code generation. Specifically, we assess performance on multi-turn dialogue generation using MT-bench~\cite{zheng2023judging}, machine translation using WMT14 DE-EN~\cite{wmt14}, summarization using CNN/Daily Mail~\cite{cnndm}, question answering using Natural Questions~\cite{nq}, mathematical reasoning using GSM8k~\cite{gsm8k}, and retrieval-augmented generation using Natural Questions with concatenated 5 Wikipedia documents~\cite{dpr}. Additionally, we evaluate code generation capabilities using HumanEval~\cite{humaneval}.

\paragraph{Models} Following SpecBench~\cite{specbench}, we employ \texttt{Vicuna-7b-v1.5}~\cite{vicuna2023} as the target model and \texttt{Vicuna-68m}~\cite{yang2024multi} as the draft model. To assess acceleration performance on base models, we also evaluate our method on \texttt{Llama2-7b}~\cite{touvron2023llama2openfoundation}.

\paragraph{Baselines} In addition to vanilla autoregressive generation, we compare DuoDecoding against four representative methods: Speculative Decoding (SpS)~\cite{leviathan2023fast}, Prompt Lookup Decoding (PLD)~\cite{pld}, Retrieval-based Speculative Decoding (REST)~\cite{REST}, and Lookahead Decoding~\cite{Lookahead}.

\paragraph{Hardware and Implementation Details} All experiments are conducted on a single A800 GPU and 16-core Intel Xeon CPU. Our implementation primarily builds on the \texttt{transformers} library~\cite{wolf2019huggingface}, with CPU inference implemented through the Python interface~\cite{llama_cpp_python} of llama.cpp~\cite{llama_cpp}. Models on GPU run in FP16 precision, while models on CPU run in the \texttt{GGUF} format with \texttt{Q5\_K\_M} quantization~\cite{ggml}.

\begin{figure*}[ht]
    \centering
    \includegraphics[width=1\linewidth]{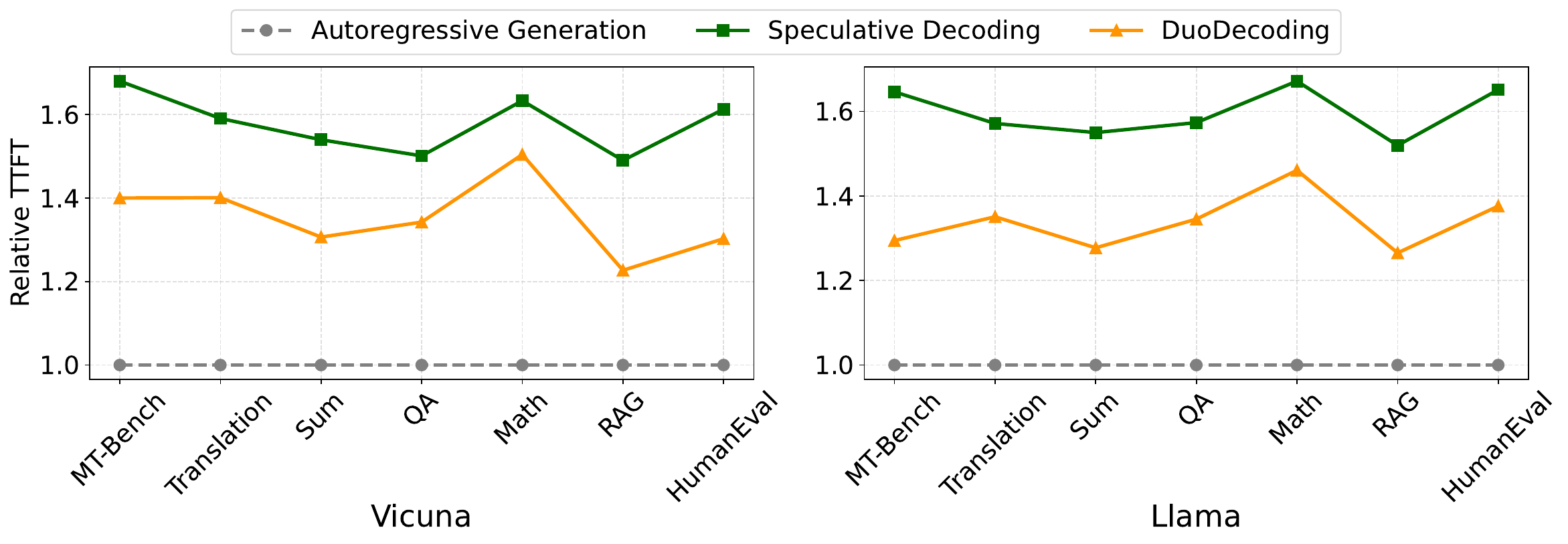}
    \caption{Comparison of Time to First Token (TTFT) across different tasks and models. The y-axis shows the relative TTFT normalized by vanilla autoregressive generation. Lower values indicate better latency performance.}
    \label{fig:ttft}
\end{figure*}

\subsection{Main Results}

Table \ref{tab:main-results} presents the comprehensive evaluation results of DuoDecoding and baseline methods. We report tokens per second (\textit{TPS}) and the speedup ratio ($\varphi$) relative to vanilla autoregreesive generation.

DuoDecoding demonstrates consistent superior performance across all tasks and model architectures. For \texttt{Vicuna-7b-v1.5}, our method achieves an average speedup of 1.58×, outperforming all baseline methods. The improvement is particularly notable in MT-bench and summarization tasks, where DuoDecoding reaches 1.67× speedup. While PLD shows strong performance in summarization (2.15×), it exhibits significant performance variations across different tasks. Other baselines like SpS and Lookahead demonstrate moderate but stable improvements, with average speedups of 1.42× and 1.35× respectively.

The advantages of DuoDecoding become even more pronounced when applied to \texttt{Llama2-7b}, achieving a remarkable average speedup of 2.57×. The method shows exceptional effectiveness in mathematical reasoning (3.31×), question answering (3.18×), and translation (3.00×) tasks. This substantial improvement over \texttt{Vicuna-7b-v1.5} suggests a higher consistency between the output distributions of draft and target models. The enhanced performance also indicates that employing a more capable draft model could potentially yield even higher speedup ratios. While baseline methods like PLD and SpS also show stronger performance on \texttt{Llama2-7b} (2.10× and 1.80× respectively), they still fall significantly short of DuoDecoding's consistent high performance across all tasks.

These results demonstrate that DuoDecoding not only provides superior acceleration but also maintains consistent performance across different tasks, addressing the stability limitations observed in existing methods.

\subsection{Time to First Token}

While overall generation latency improvements are crucial, the latency of producing the first token (TTFT) is equally important for real-world applications, especially in interactive scenarios. We compares the relative TTFT of DuoDecoding against Speculative Decoding across different tasks in Figure~\ref{fig:ttft}.

Both \texttt{Vicuna-7b-v1.5} and \texttt{Llama2-7b} exhibit similar trends across different tasks, demonstrating the consistent behavior of these acceleration methods. DuoDecoding maintains lower TTFT overhead compared to Speculative Decoding across all tasks, with an average relative TTFT around 1.3-1.4×. In contrast, Speculative Decoding shows higher latency overhead with relative TTFT ranging from 1.5× to 1.7×. On average, DuoDecoding's TTFT is approximately 83\% of Speculative Decoding's TTFT, representing a significant improvement in initial response time.

The superior TTFT performance of DuoDecoding stems from its efficient utilization of heterogeneous computing resources, allowing parallel forward passes of both draft and target models. In contrast, Speculative Decoding requires sequential execution of these operations. However, it's worth noting that DuoDecoding still exhibits higher TTFT compared to vanilla autoregressive generation, primarily due to the additional system overhead and verification operations required before generating the first token. Moreover, the draft model may take longer than the target model during parallel decoding.

\subsection{Analysis}
\subsubsection{Ablation Study}
We conduct ablation experiments using \texttt{Llama-2-7b} on multi-turn dialogue, translation, mathematical reasoning, and code generation tasks, and report the number of tokens generated per second (TPS).

\paragraph{Dynamic Multi-Sequence Drafting}

\begin{table}[t]
\centering
\begin{tabular}{@{}llcccc}
\toprule
\multicolumn{2}{l}{\textbf{\#Seq}}                              & \textbf{MT} & \textbf{Trans} & \textbf{Math} & \textbf{Code} \\ \midrule
                                                            & 1 & 98.71       & 131.15         & 149.68        & 89.06         \\
                                                            & 2 & 102.30       & 118.48         & 136.00           & 84.15         \\
\multirow{-3}{*}{\rotatebox{90}{Static}} & 3 & 101.22      & 113.42         & 124.76        & 80.54         \\
\rowcolor[HTML]{e0e0e0} 
\multicolumn{2}{l}{\cellcolor[HTML]{e0e0e0}Dynamic} & \textbf{101.67} & \textbf{139.08} & \textbf{150.67} & \textbf{89.52} \\ \bottomrule
\end{tabular}
\caption{Impact of different sequence drafting strategies.}
\label{tab:dynamic-seq}
\end{table}

Table~\ref{tab:dynamic-seq} presents a comparative analysis of different sequence drafting strategies.
Our dynamic sequence drafting method consistently outperforms other approaches across all four tasks, highlighting its effectiveness.

For static sequence numbers, no single configuration emerges as universally optimal across all tasks. The sequence number of 1 generally performs well, due to its ability to maximize draft length under the fixed computational budget. Increasing the sequence number will reduce the available draft length for each sequence.
However, MT-bench presents an exception where sequence numbers greater than 1 show better performance. This can be attributed to cases where the draft model's initial predictions are incorrect. In such scenarios, multiple shorter sequences are more advantageous than a single longer sequence, as they allow for more diverse speculation paths.

These findings support the necessity of dynamically adjusting the sequence number. Different contexts and generation stages may benefit from varying sequence numbers, and our dynamic approach successfully adapts to these changing requirements, leading to the best performance across diverse tasks among these sequence drafting strategies.

\paragraph{Hardware-aware Drafting Budget}

\begin{table}[t]
\centering
\begin{tabular}{ccccc}
\toprule
\textbf{Budget} &\textbf{MT} & \textbf{Trans} & \textbf{Math} & \textbf{Code} \\ \midrule
$\gamma-2$ & 97.44          & 128.65          & 145.32          & 87.47          \\
$\gamma-1$ & 97.77          & 130.75          & 147.90          & 89.00          \\
\rowcolor[HTML]{e0e0e0} 
$\gamma$   & \textbf{98.71} & \textbf{131.15} & \textbf{149.68} & \textbf{89.06} \\
$\gamma+1$ & 98.08          & 130.90          & 147.12          & 88.00          \\
$\gamma+2$ & 98.38          & 130.37          & 148.60          & 87.11          \\ \bottomrule
\end{tabular}
\caption{Impact of different drafting budgets. $\gamma$ represents the optimal hardware-aware budget.}
\label{tab:budget}
\end{table}

Table~\ref{tab:budget} investigates the effect of varying the drafting budget around our hardware-aware optimal value, $\gamma$. To isolate the impact of drafting budget adjustments and eliminate confounding factors, we fix the number of draft sequences to 1 across all experiments. The hardware-aware budget $\gamma$ consistently delivers the best performance across all tasks.

When the budget is lower than $\gamma$, the draft model finishes generation early and remains idle, missing the opportunity to generate more tokens. Conversely, when the budget exceeds $\gamma$, the target model experiences idle time while waiting for the draft model to complete its generation. In both cases, suboptimal resource utilization leads to decreased performance.

While the performance differences between various budget settings are relatively slight, this is primarily due to our baseline $\gamma$ being set to 24 tokens. Variations of ±1 or ±2 tokens represent small proportional changes relative to this substantial base value.

\subsubsection{Profiling}

\begin{figure}[t]
    \centering
    \includegraphics[width=1\linewidth]{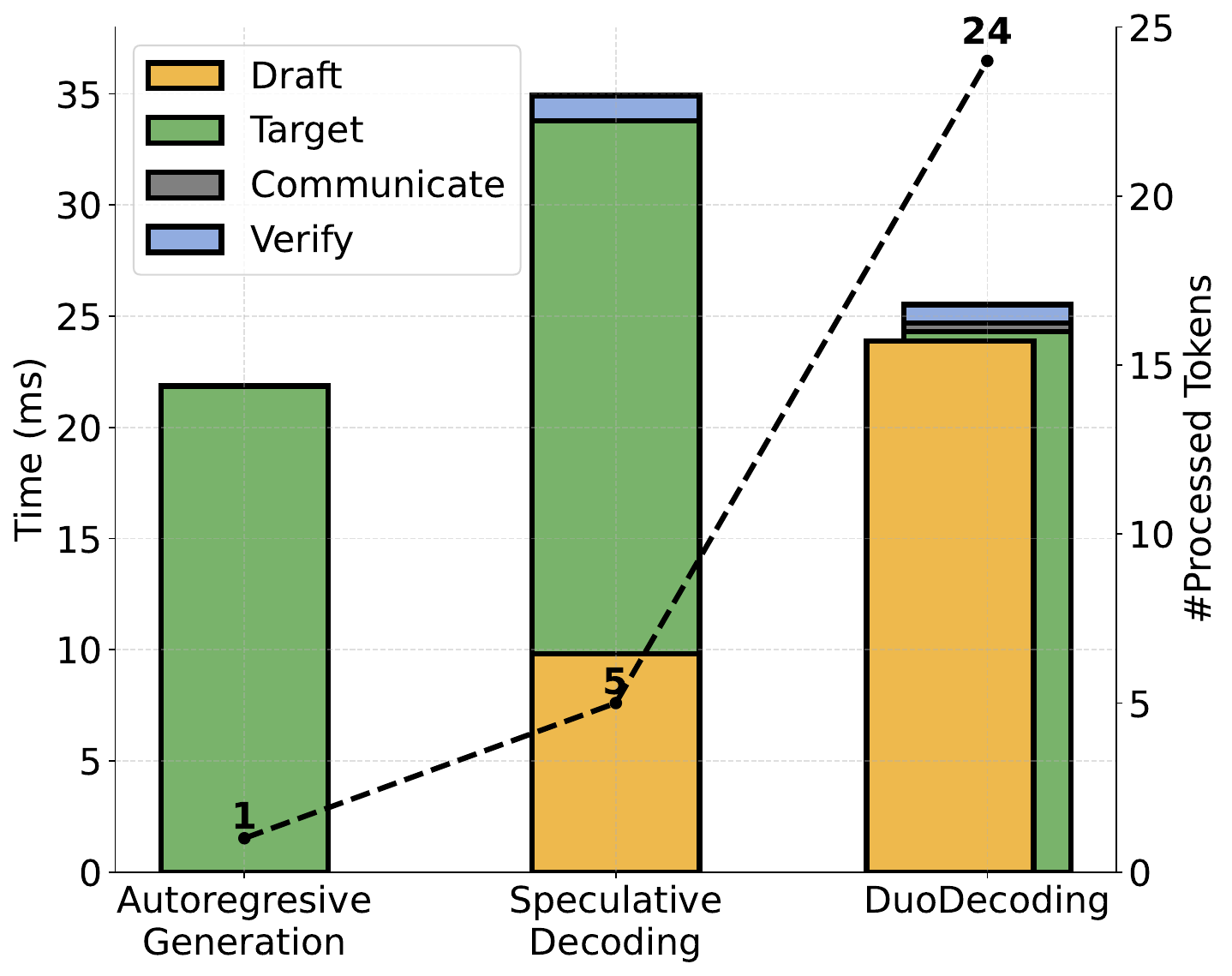}
    \caption{Profiling of time and number of processed token in one generation iteration for different decoding strategies.}
    \label{fig:profiling}
\end{figure}

We provide a detailed profiling of the time and number of processed tokens per iteration for different decoding strategies in Figure~\ref{fig:profiling}.

Autoregressive generation shows the lowest iteration time but processes only one token per iteration. This represents the baseline approach with minimal overhead but high average latency.

Speculative decoding demonstrates significantly higher iteration time, with a substantial portion consumed by the draft model. While this increases the total processing time, it enables the handling of 5 tokens per iteration, significantly reducing the average latency. The verification and communication overhead are negligible.

DuoDecoding improves efficiency by parallel execution of draft and target models. While maintaining a similar iteration time to autoregressive generation, it dramatically increases the number of processed tokens to 24 per iteration. The draft model execution overlaps with the target model's computation, effectively eliminating the additional time overhead seen in speculative decoding. As with speculative decoding, the verification and communication costs remain minimal.

\subsubsection{Sequence Number Prediction}

\begin{table}[t]
\centering
\begin{tabular}{|c|c|c|c|}
\hline
\diagbox{Actual}{Pred} & =1 & >1 & Total \\
\hline
=1 & 0.348 & 0.087 & 0.435 \\ \hline
>1 & 0.230 & 0.335 & 0.565 \\
\hline
Total & 0.578 & 0.422 & 1.000 \\
\hline
\end{tabular}
\caption{Confusion matrix of acutal and predicted sequence number.}
\label{tab:confusion-matrix}
\end{table}

In Table~\ref{tab:confusion-matrix}, we analyze the prediction accuracy of our dynamic sequence drafting strategy. 
Specifically, "Actual" refers to whether the top-1 probability draft sequence in the decoding process is accepted, while "Pred" indicates whether the number of sequences used in DuoDecoding exceeds 1.
Since we compare our dynamic multi-sequence drafting strategy against static single-sequence drafting (sequence number = 1), we specifically focus on cases where predicted and actual sequence numbers are equal to or greater than 1.

The analysis shows that in 56.5\% of cases where the sequence number is 1, the entire prediction sequence would be rejected. 
This indicates a significant proportion of cases where speculative prediction requires drafting more diverse tokens starting from the first position to improve the acceptance rate.
For predictions of sequence numbers greater than 1, we achieved good prediction accuracy: only 8.7\% of cases were incorrectly predicted (i.e., should have been 1), while in 33.5\% of cases, we accurately predicted the opportunity to use multiple sequences, allowing us to explore more tokens. This low false positive rate not only effectively reduces computational resource waste but also provides substantial opportunities for acceleration.

Together with our previous experimental analysis, our dynamic multi-sequence drafting strategy effectively balances accuracy and efficiency, successfully improving overall performance.

\subsubsection{Sequence Number Distribution}
\begin{figure}
    \centering
    \includegraphics[width=1\linewidth]{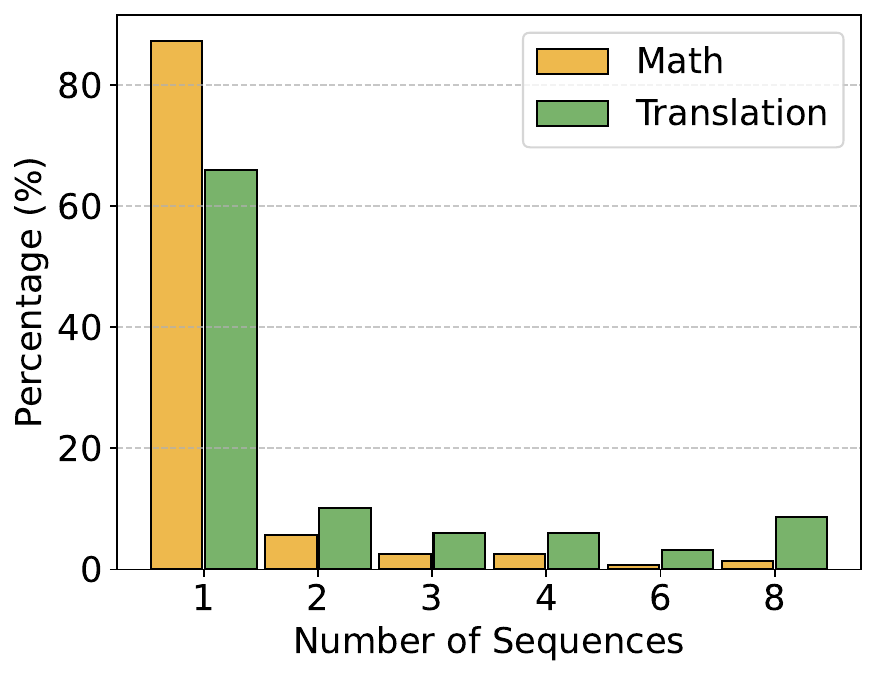}
    \caption{Distribution of sequence numbers during generation process in DuoDecoding.}
    \label{fig:seq-num}
\end{figure}

We evaluated the distribution of the number of sequences during actual execution on two tasks,Math and Translation, in Figure~\ref{fig:seq-num}. In both tasks, the case where the sequence number is 1 occurs much more frequently than other sequence numbers.

The distribution of sequence numbers differs significantly between the two tasks. In the Math task, the majority of sequences have a sequence number of 1, while in the Translation task, this occurs much less frequently, with a larger proportion of cases using higher sequence numbers. 
We attribute this difference to the generation process of the two tasks. In the Math task, many sequences involve the repetition of numbers and previously mentioned names or entities, allowing the draft model to predict most tokens with high probability. In contrast, the Translation task involves more distinct input-output pairs, with translations of the same sentence showing greater diversity. This increased variability leads to higher uncertainty, necessitating the use of more sequences.

\section{Conclusion}
In this work, we propose DuoDecoding, a heterogeneous speculative decoding method designed to reduce the draft overhead inherent in conventional speculative decoding. 
We strategically deploy the draft and target models on CPU and GPU respectively, enabling parallel execution. 
We minimize the idle time caused by mutual waiting between the CPU and GPU with hardware-aware optimal drafting budget.
Additionally, we propose to draft with dynamic multi-sequences to enhance the quality of the draft. 
Extensive experiments across multiple tasks demonstrate that DuoDecoding consistently achieves lower generation latency compared to baseline methods. Further ablation and analysis confirm the effectiveness of each component of our approach. 

We hope that our work will inspire further research on leveraging heterogeneous resources for language model inference, as the speculative decoding framework provides a promising solution for collaborative inference.

\section*{Limitations}
Our work has several limitations. First, since speculative decoding primarily focuses on reducing generation latency, we did not explore the performance of different methods under large batch sizes. Second, although we tested the performance across both base and chat models, our experiments were limited to target models with 7B parameters, and the effectiveness of our approach on larger models remains unexplored. Finally, due to hardware constraints, our evaluations were conducted on a single hardware configuration, and the performance characteristics on different computing platforms remain to be investigated.

\bibliography{custom}

\newpage
\appendix

\section{Experimental Details}
\paragraph{Template} For \texttt{Vicuna-7B-v1.5}, we used the official template. For \texttt{Llama-2-7B}, the templates we used are as follows.
\begin{tcolorbox}[title=MT-Bench]
A chat between a curious user and an artificial intelligence assistant. The assistant gives helpful, detailed, and polite answers to the user's questions. USER: \{\{QUESTION\}\} ASSISTANT:
\end{tcolorbox}
\begin{tcolorbox}[title=Translation]
Translate German to English. German: \{\{QUESTION\}\} English:
\end{tcolorbox}
\begin{tcolorbox}[title=Summarization]
Summarize:

\{\{QUESTION\}\}

TL;DR:
\end{tcolorbox}
\begin{tcolorbox}[title=QA]
A chat between a curious user and an artificial intelligence assistant. The assistant gives helpful, detailed, and polite answers to the user's questions. USER: \{\{QUESTION\}\} ASSISTANT:
\end{tcolorbox}
\begin{tcolorbox}[title=Math]
\{\{QUESTION\}\} Let's think step by step.
\end{tcolorbox}
\begin{tcolorbox}[title=RAG]
A chat between a curious user and an artificial intelligence assistant. The assistant gives helpful, detailed, and polite answers to the user's questions. USER: \{\{QUESTION\}\} ASSISTANT:
\end{tcolorbox}
\begin{tcolorbox}[title=Code]
\{\{QUESTION\}\} 
\end{tcolorbox}


\paragraph{Datasets} The datasets included in SpecBench~\cite{specbench} are the same 80 samples as those used in SpecBench. For Humaneval~\cite{humaneval}, we use the full set of 164 samples.

\section{License for Scientific Artifacts}
In this research, we strictly adhere to the license terms of all utilized datasets and models. All resources are publicly available and our usage complies with their original intended purposes and license scopes.

\end{document}